\documentclass[11pt]{article}

\usepackage[margin=1in]{geometry}

\usepackage[utf8]{inputenc}
\usepackage[T1]{fontenc}
\usepackage{lmodern}
\usepackage{hyperref}
\usepackage{url}
\usepackage{booktabs}
\usepackage{amsfonts}
\usepackage{amsmath}
\usepackage{amssymb}
\usepackage{amsthm}
\usepackage{nicefrac}
\usepackage{microtype}
\usepackage{xcolor}
\usepackage{graphicx}
\usepackage{enumitem}
\usepackage{natbib}
\usepackage{authblk}

\usepackage{tikz}
\usetikzlibrary{positioning,arrows.meta,shapes.multipart}

\definecolor{labeled}{RGB}{255, 235, 130}
\definecolor{unlabeled}{RGB}{170, 240, 170}
\definecolor{module}{RGB}{240,240,240}

\hypersetup{
    colorlinks=true,
    linkcolor=black,
    citecolor=blue,
    urlcolor=blue
}

\title{In-Context Positive-Unlabeled Learning}

\author{Siyan Liu}
\author{Yi Chang}
\author{Manli Cheng}
\author{Qinglong Tian}
\author{Pengfei Li}
\affil{Department of Statistics and Actuarial Science, University of Waterloo}
\date{}

\begin{document}

\maketitle

\begin{abstract}
Positive-unlabeled (PU) learning addresses binary classification when only a set of labeled positives is available alongside a pool of
unlabeled samples drawn from a mixture of positives and negatives. Existing
PU methods typically require dataset-specific training or iterative
optimization, which limits their applicability when many tasks must be solved
quickly or with little tuning. We introduce \textbf{PUICL}, a pretrained
transformer that solves PU classification entirely through in-context
learning. PUICL is pretrained on synthetic PU datasets generated from
randomly instantiated structural causal models, exposing it to a wide range
of feature-label relationships and class-prior configurations. At inference
time, PUICL receives the labeled positives and the unlabeled samples as a
single input and returns class probabilities for the unlabeled rows in one
forward pass, with no gradient updates or per-task fitting. On 20
semi-synthetic PU benchmarks derived from the UCI Machine Learning
Repository, OpenML, and scikit-learn, PUICL outperforms four standard PU
learning baselines in average AUC and accuracy, and is competitive on
$F_1$-score. These results show that the in-context learning paradigm
extends naturally beyond fully supervised tabular prediction to the
semi-supervised PU setting.

\end{abstract}

\section{Introduction}

\subsection{Positive and Unlabeled Data}

Positive and unlabeled (PU) data arise when only a subset of positive examples is labeled, while the remaining data consist of a mixture of positive and negative instances. Such settings are common in many applications, including medical research, ecology, and machine learning. For example, in contaminated case-control studies, diagnosed patients form a labeled positive group, while the control group, which is intended to contain healthy individuals, may include undiagnosed cases, resulting in contamination. Similarly, in ecological studies, presence-only data record observed occurrences of a species, while unobserved locations remain unlabeled rather than truly negative. Analogous structures also appear in applications such as text classification, recommendation systems, and medical screening, where positive labels can be obtained with high confidence but reliable negative labels are difficult or costly to acquire.

A standard assumption in PU learning is the \emph{selected completely at random} (SCAR) assumption, which posits that the conditional feature distribution of the positive class is identical for labeled and unlabeled positives. Under SCAR, the observed data follow
\[
\begin{split}
\mathbf{x}_{1},\dots,\mathbf{x}_{n} &\sim f_{+}(\mathbf{x}),\\
\mathbf{x}_{n+1},\dots,\mathbf{x}_{n+m} &\sim (1-\pi)\, f_{+}(\mathbf{x}) + \pi\, f_{-}(\mathbf{x}),
\end{split}
\]
where \(\pi \in (0,1)\) denotes the negative prevalence in the unlabeled pool, and \(f_{+}(\mathbf{x})\) and \(f_{-}(\mathbf{x})\) are the feature distributions of the positive and negative classes, respectively. In practice, all three quantities are unknown.

The goal of PU learning is to construct a classifier that distinguishes positive from negative instances using only PU data. This problem is inherently more challenging than standard supervised classification due to the absence of fully observed negative samples. As a result, successful methods must exploit the mixture structure of the unlabeled data together with the information provided by labeled positives, making PU learning a distinctive form of semi-supervised learning.

\subsection{In-context Learning}

Modern large language models (LLMs), such as ChatGPT, do not retrain their parameters for each new task. Instead, they adapt their predictions based on examples or instructions provided in the input, all within a single forward pass. This phenomenon is known as \emph{in-context learning} (ICL). More broadly, ICL refers to a paradigm in which a pretrained model adapts to a new prediction task using the data supplied at inference time, without any parameter updates.

From a statistical perspective, ICL suggests an alternative mode of learning. Rather than fitting a new model separately for each dataset, as in conventional supervised learning, one can instead train a general-purpose prediction mechanism that learns to condition on data and perform inference directly. This paradigm has recently shown strong empirical success across a variety of domains.

\subsection{Related Work}

\paragraph{PU Learning}

Most existing work on PU learning has focused on prediction, particularly on training classifiers from data containing only positive and unlabeled examples. Early work by \citet{elkan_learning_2008} introduced a probabilistic formulation of PU classification, which led to a large body of classifier-oriented approaches. These include risk-reweighting and unbiased risk estimation methods \citep{plessis2014analysis, kiryo_positive-unlabeled_2017}, approaches that incorporate additional structural assumptions such as known class priors \citep{hido_statistical_2011, kato_learning_2018}, and methods based on the anchor set assumption and its variants \citep{chen_variational_2020, garg_mixture_2021}. Despite their differences, these methods all operate within the standard (semi-)supervised learning paradigm, where a model is explicitly trained or optimized for each dataset.

\paragraph{Tabular Foundation Models}

Recent work on tabular foundation models has shifted the paradigm of tabular learning from dataset-specific training to in-context learning. In this framework, a pretrained model receives the training data as context and produces predictions for new observations in a single forward pass. A representative example is TabPFN \citep{hollmann2025accurate}, which pretrains a transformer on synthetic prior data and demonstrates strong performance on small tabular classification tasks without per-dataset optimization.

Beyond TabPFN, several recent models have expanded this paradigm toward improved scalability and realism. For instance, TabICL \citep{qu2025tabicl} introduces a two-stage architecture that enables efficient in-context learning on substantially larger datasets, while TabDPT \citep{ma2025tabdpt} emphasizes pretraining on real tabular data and exhibits favorable scaling behavior with respect to both model and data size. These developments suggest that tabular foundation models are evolving into a general-purpose framework for tabular prediction.

\subsection{Our Contributions}

We introduce \textbf{PUICL}, a pretrained transformer model for PU learning that operates entirely through in-context learning. Unlike existing PU methods, which require dataset-specific model fitting or iterative optimization, PUICL performs classification in a single forward pass without updating model parameters. This represents a fundamental shift in how PU learning problems can be approached, replacing per-dataset training with a general-purpose inference mechanism learned through pretraining.

From the perspective of tabular foundation models, our work extends their applicability beyond standard supervised learning to the semi-supervised setting. While existing models such as TabPFN focus primarily on supervised prediction, PUICL demonstrates that the in-context learning paradigm can be successfully adapted to problems with incomplete labeling, thereby broadening the scope of tabular foundation models.

On the methodological side, we develop a prior data generation procedure specifically tailored to PU learning, enabling effective pretraining under the in-context framework. The model adopts a TabPFN-style transformer architecture and is trained using a curriculum designed to improve robustness across diverse data regimes. To facilitate reproducibility and future research, we make the entire pipeline fully open-source, including the prior data generation process, training curriculum, and model implementation.

The rest of this paper is organized as follows.
Section~\ref{sec:synthetic-data} describes the prior data generation procedure based on randomly instantiated structural causal models, together with the construction of synthetic PU training instances.
Section~\ref{sec:architecture} presents the PUICL architecture and explains how the input encoding, dual-axis transformer, and output decoder are adapted to the PU setting.
Section~\ref{sec:pretraining} details the pretraining curriculum and the in-context inference procedure.
Section~\ref{sec:experiments} reports empirical results on 20 semi-synthetic PU classification tasks and compares PUICL against four standard PU learning baselines.
Section~\ref{sec:discussion} concludes with a discussion of limitations and future directions.

\section{Prior Data Generation}
\label{sec:synthetic-data}

To pretrain the model in-context, we require a distribution over labeled datasets
that captures the diversity of real-world PU learning problems.
We construct this prior using randomly instantiated structural causal models (SCMs),
following the meta-learning prior philosophy of \citet{hollmann2025accurate, qu2025tabicl}. Each
training step draws a fresh synthetic dataset from this prior, exposing the model to
a wide variety of data-generating mechanisms without any real labeled data.

\subsection{SCM Construction.}
Each dataset is generated by a distinct SCM, instantiated as a randomly initialized
multilayer perceptron (MLP). Concretely, the SCM consists of $k$ exogenous (root)
variables, a depth-$L_g$ MLP with hidden dimension $h$, and two readout heads for
features and labels respectively. The MLP weights are drawn independently from
$\mathcal{N}(0, \sigma_{\mathrm{init}}^2)$ with zero biases, and the activation
function of each hidden layer is drawn uniformly from a fixed bank of ten
nonlinearities: $\tanh$, ReLU, GELU, the identity $\phi(x) = x$, the sign
function $\phi(x) = \mathrm{sign}(x) \in \{-1, 0, 1\}$, the Heaviside step
function $\phi(x) = \mathbf{1}[x > 0]$, a radial basis function
$\phi(x) = \exp(-x^2)$, the sine function $\phi(x) = \sin(x)$, the square
function $\phi(x) = x^2$, and the absolute value $\phi(x) = |x|$. Because a
new MLP is instantiated for every dataset, the collection of all such MLPs
implicitly defines a rich nonparametric prior over functional relationships between
inputs and outputs.

\subsection{Forward Pass, Feature and Label Generation}
To generate $n$ samples, we first draw exogenous cause vectors
$\mathbf{u}_i \in \mathbb{R}^k$ i.i.d.\ from $\mathcal{N}(\mathbf{0}, \mathbf{I})$
(or $\mathrm{Uniform}(0,1)^k$, depending on the configuration). These are propagated
through the MLP as follows. Let $\mathbf{h}_i^{(0)} = \mathbf{W}_0\,\mathbf{u}_i$
be the output of the input linear layer. For each subsequent layer
$\ell = 1, \ldots, L_g-1$:
\[
    \mathbf{h}_i^{(\ell)}
    \;=\;
    \mathbf{W}_\ell\,\phi_\ell\!\left(\mathbf{h}_i^{(\ell-1)}\right)
    + \boldsymbol{\varepsilon}_i^{(\ell)},
    \qquad
    \boldsymbol{\varepsilon}_i^{(\ell)} \sim \mathcal{N}\!\left(\mathbf{0},\,
    \sigma_{\mathrm{noise}}^2\,\mathbf{I}\right),
\]
where $\phi_\ell$ is the activation function of layer $\ell$. The per-layer
Gaussian noise $\boldsymbol{\varepsilon}_i^{(\ell)}$ introduces stochasticity into
the functional relationships, reflecting the fact that real SCMs are rarely
deterministic.

How the feature vector $\mathbf{x}_i \in \mathbb{R}^d$ and the latent label score
$s_i \in \mathbb{R}$ are jointly extracted depends on the chosen generation mode.
In the \emph{non-causal} mode, both quantities are read off from the final hidden
state $\mathbf{h}_i^{(L_g-1)}$ via dedicated linear readout heads: a feature
readout $\mathbf{R}_x \in \mathbb{R}^{d \times h}$ yields
$\mathbf{x}_i = \mathbf{R}_x\,\mathbf{h}_i^{(L_g-1)}$, and a label
readout $\mathbf{r}_y \in \mathbb{R}^h$ yields $s_i = \mathbf{r}_y^\top
\mathbf{h}_i^{(L_g-1)}$. Alternatively, the features can be set directly to the
exogenous causes, $\mathbf{x}_i = \mathbf{u}_i$ (requiring $d = k$), while the
label score is still produced by the label readout; this bypasses the MLP
nonlinearities for the features entirely.
In the \emph{causal} mode, no readout heads are used. Instead, all intermediate
hidden states $\mathbf{h}_i^{(1)}, \ldots, \mathbf{h}_i^{(L_g-1)}$ are concatenated
into a single pool of $(L_g-1)h$ latent variables, each corresponding to an endogenous
node in the SCM. One node is selected uniformly at random to serve as the label
variable $s_i$, restricted to either the first hidden block (placing the label at a
causally upstream position) or the last hidden block (placing it at a causally
downstream position). The $d$ feature variables $\mathbf{x}_i$ are then selected
from the remaining nodes, either as a contiguous clique surrounding the label node
(to preserve dense local causal dependencies) or as a random subset drawn without
replacement from the full pool. This construction ensures that features and
label are genuine intermediate nodes of the SCM, with their relationships governed
entirely by the network's weight matrices and activation functions.

In all modes, each feature dimension and the score are independently standardized to
zero mean and unit variance across the $n$ samples and clipped to $[-20, 20]$ for
numerical stability. Binary labels are then assigned by ranking samples on $s_i$:
the top $\lceil \pi n \rceil$ samples are designated as \emph{negatives} ($y_i = -$),
while the remaining samples are designated as \emph{positives} ($y_i = +$), where
$\pi \in (0,1)$ is the target negative prevalence.
This rank-based thresholding ensures that the label is a deterministic, monotone
function of the latent score, with classification difficulty governed entirely by
the geometry of the SCM.

\subsection{Constructing the PU Training Set}
The fully labeled dataset is converted into a PU learning instance by the following
strategy. We generate one binary-labeled dataset of size $n = n_{\mathrm{tr}} +
n_u$ and split it into two portions: a \emph{pre-removal training
portion} of size $n_{\mathrm{tr}}$ and an \emph{unlabeled portion} of size $n_u$,
both containing positive ($y=+$) and negative ($y=-$) samples. We then remove all
negatives from the training portion, leaving only $P$ positives, which serve as the
labeled positive data. The unlabeled portion has its labels hidden and is presented to the
model as unlabeled data. Combining these two components yields a PU dataset that
mimics realistic PU learning scenarios, where only a small set of confirmed positives
is available alongside a large pool of unlabeled data of unknown composition.

This process is governed by three parameters: the number of labeled positives $P$,
the unlabeled-to-positive ratio $\eta > 0$, and the negative prevalence $\pi \in
(0, 1)$ shared by both the training portion and the unlabeled set. Since the
pre-removal training portion contains $P$ positives at a positive prevalence of
$1-\pi$, its size is $n_{\mathrm{tr}} = \lceil P / (1 - \pi) \rceil$, of which
$\pi \cdot n_{\mathrm{tr}}$ negatives are removed after the split. The unlabeled
set size is $n_u = \lceil P \cdot \eta \rceil$. To form the split with
controlled class composition, the positive and negative samples are partitioned into
separate pools, shuffled independently, and allocated to each portion such that the
realized negative prevalence in both portions approximates $\pi$. The final PU
dataset presented to the model consists of $P$ labeled positives and $n_u$
unlabeled samples whose labels are hidden during training and revealed only for
evaluation.

\section{Model Architecture}
\label{sec:architecture}
\begin{figure}[ht]
    \centering
    \includegraphics[width=1\linewidth]{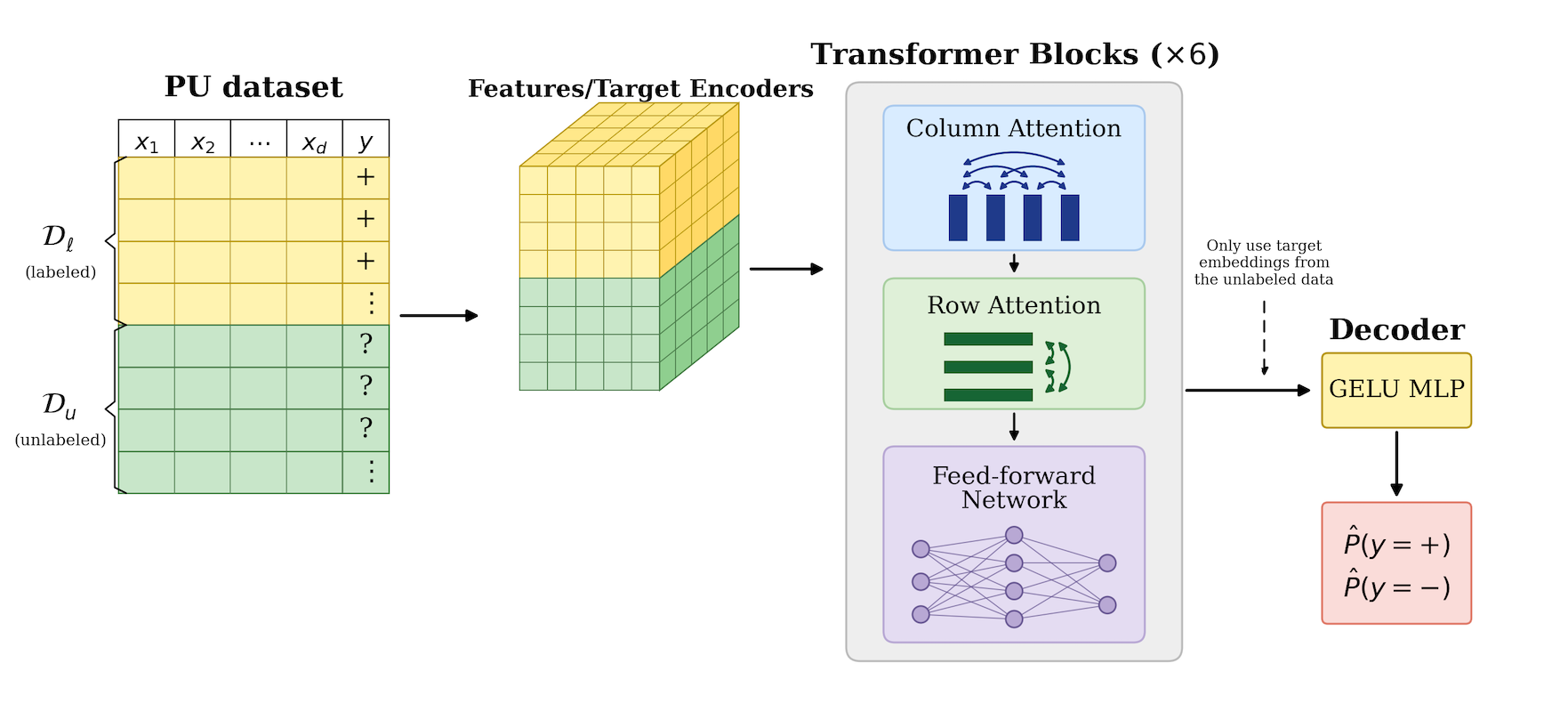}
    \caption{Architecture of the in-context PU learning model. The input PU dataset
    $\mathcal{D}_\ell \cup \mathcal{D}_u$, where $\mathcal{D}_\ell$ denotes the
    set of $P$ labeled positives and $\mathcal{D}_u$ the set of $n_u$ unlabeled
    samples, is encoded into a 4-D embedding tensor of shape
    $B \times n \times (d+1) \times e$ (batch size, rows, columns, embedding
    dimension), processed by $L$ transformer blocks each applying column attention
    (across features), row attention (across data points), and a feed-forward
    layer, and finally decoded into class probabilities $\hat{P}(y{=}+)$ and
    $\hat{P}(y{=}-)$ for each unlabeled sample via a GELU MLP applied to the
    unlabeled-row target embeddings.}
    \label{fig:model_arch}
\end{figure}

We adopt a tabular transformer architecture modified from \citet{hollmann2025accurate, pfefferle2025nanotabpfn}
for the PU learning setting. The model takes a PU dataset as input, which consists of
$P$ labeled positives and $n_u$ unlabeled samples, each with $d$
features, and outputs a class probability for each unlabeled sample, all in a
single forward pass without any gradient update at inference time.
Figure~\ref{fig:model_arch} illustrates the flow of a PU dataset through the model architecture.

\subsection{Input Encoding}

The input dataset is represented as a matrix of $n_u + P$ rows and $d+1$
columns, where the extra column holds the label. Each entry of this matrix is a
scalar, and the model embeds every scalar independently into a vector of dimension
$e$ (the embedding size) before any attention is applied.

\paragraph{Feature encoder.}
Each feature value $x_{ij}$ (the value of feature $j$ for sample $i$) is encoded by
a single shared linear layer $\mathbf{W}_x \in \mathbb{R}^{1 \times e}$. Before
projection, each feature column is standardized using the mean and standard deviation
computed from the labeled positive rows only, and clipped to $[-100, 100]$.

\paragraph{Target encoder.}
The label column requires special treatment because unlabeled samples have no
observed label. For labeled positives, the label scalar is projected by a linear
layer $\mathbf{w}_y \in \mathbb{R}^{1 \times e}$, identical in form to the feature
encoder. For unlabeled samples, a dedicated encoding is needed.

Models like TabPFN \citep{hollmann2025accurate} handle unobserved test labels by
imputing them with the mean of the training labels. In standard supervised
in-context learning this is benign.
In the PU
setting, however, all labeled training examples are positives, so the mean label is
a constant. Imputing every unlabeled entry with this constant would assign them all
the same positive-class encoding, making the model unable to distinguish positives
from negatives among the unlabeled samples, which will result in a fundamental identifiability failure.

We therefore replace mean imputation with a learnable \emph{unlabeled token}
$\mathbf{v} \in \mathbb{R}^e$, a globally shared parameter trained jointly with the
rest of the model. This token is used as the label embedding for every unlabeled
sample, regardless of its true (hidden) class. Rather than injecting a fixed,
potentially misleading signal, the unlabeled token allows the model to learn, through
pretraining, the most informative representation for an entry whose label is unknown
in a PU context.
\paragraph{Input fusion.}
The $d$ feature embeddings and the single label embedding for each row are
concatenated along the column dimension, yielding a 4-D representation
$\mathbf{H}^{(0)} \in \mathbb{R}^{B \times n \times (d+1) \times e}$, where $B$ is
the batch size and $n = P + n_u$ is the total number of rows. This
representation can be thought of as a table of embedding vectors, with one vector
per cell.

\subsection{Dual-Axis Transformer}

The core of the model is a stack of $L$ transformer encoder blocks, each of which
applies two successive multi-head self-attention operations along orthogonal axes of
the table, followed by a position-wise feed-forward network.

\paragraph{Within-row attention (between features).}
The first attention operation is applied independently to each row. The table is
reshaped to $(B \cdot n) \times (d+1) \times e$, so that each row becomes a
sequence of $d+1$ token embeddings (one per feature plus the label). Multi-head
self-attention is applied across these tokens, allowing the model to capture
interactions among features and between features and the label within a single
sample. The output is reshaped back to the full table and passed through a layer
normalization.

\paragraph{Within-column attention (between data points).}
The second attention operation is applied independently to each column. The table is
transposed and reshaped to $(B \cdot (d+1)) \times n \times e$, so that each column
becomes a sequence of $n$ row embeddings. Multi-head self-attention is then applied
across all $n$ rows without any masking, enabling every sample, either labeled or
unlabeled, to attend to every other sample in the dataset.

This design departs from the attention masking scheme used in models like TabPFN
\citep{hollmann2025accurate}. In standard supervised in-context learners, test
samples are treated as \emph{external} queries: training-set representations are
computed independently of the test inputs, and test samples are prevented from
attending to one another, since each test prediction should be conditionally
independent given the training context. Such masking reflects the assumption that
the test set plays no role in defining the learned function.

PU learning, however, is fundamentally a semi-supervised problem. The unlabeled
samples are not external queries but an integral part of the learning problem: the
global structure of the unlabeled pool carries information that is directly relevant to identifying positives within it.
Predictions are therefore \emph{within-dataset} rather than out-of-sample, and
unlabeled samples should be allowed to inform one another. We accordingly remove all
attention masks and allow unrestricted full self-attention across all rows, so that
labeled and unlabeled samples jointly shape each other's representations throughout
every layer of the network.

\paragraph{Feed-forward sublayer.}
Each block concludes with a standard position-wise feed-forward network applied to
every cell embedding:
\[
    \mathbf{H} \;\leftarrow\; \mathbf{W}_2\,\mathrm{GELU}(\mathbf{W}_1\,\mathbf{H})
    + \mathbf{H},
\]
where $\mathbf{W}_1 \in \mathbb{R}^{e \times e_{\mathrm{ff}}}$ and
$\mathbf{W}_2 \in \mathbb{R}^{e_{\mathrm{ff}} \times e}$, followed by layer
normalization.

\subsection{Output Decoder}

After $L$ transformer blocks, predictions are read from the label-column embeddings
of the unlabeled rows only. Formally, for each unlabeled row $i$, the model extracts
the embedding at the label-column position, $\mathbf{z}_i^{(L)} \in \mathbb{R}^e$,
and passes it through a two-layer MLP decoder:
\[
    \hat{\mathbf{p}}_i \;=\; \mathbf{W}_4\,\mathrm{GELU}(\mathbf{W}_3\,
    \mathbf{z}_i^{(L)}),
\]
where $\hat{\mathbf{p}}_i \in \mathbb{R}^2$ are the logits for the positive ($y{=}+$) and
negative ($y{=}-$) classes. The model is trained with a cross-entropy loss computed over the
unlabeled rows, whose true labels are available from the synthetic prior but
withheld from the model during the forward pass.

\subsection{Model Configuration}

The default model configuration uses an embedding size of $e = 128$, $L = 6$
transformer blocks, $8$ attention heads, and a feed-forward hidden size of
$e_{\mathrm{ff}} = 256$. The model has no positional encoding, consistent with the
permutation-invariant nature of tabular data: predictions should not depend on the
order in which samples or features are presented. Under this configuration, the model
has approximately $1.23$ million trainable parameters in total, which is several orders of
magnitude smaller than typical large language models, and comparable in scale to a
small convolutional network. The bulk of these parameters ($1.19$M, or $97\%$)
reside in the six transformer blocks; the feature and target encoders together
contribute fewer than $650$ parameters, and the output decoder contributes
$33{,}538$. This extreme compactness is a direct consequence of the in-context
learning paradigm: rather than encoding task-specific knowledge in a large number of
weights, the model is trained to read and reason over the dataset presented at
inference time, delegating the burden of adaptation entirely to the attention
mechanism. The result is a lightweight predictor that generalizes across PU learning
problems without any per-task fine-tuning.

\section{Pretraining and Inference}
\label{sec:pretraining}

\subsection{Pretraining}

The model is trained entirely on synthetic PU datasets sampled from the SCM prior
described in Section~\ref{sec:synthetic-data}; no real labeled data are used at any point
during training.
Each training step draws a fresh batch of $24$ independent datasets per GPU
(effective batch size $48$ across two GPUs), computes predictions for the unlabeled
rows, and minimizes the cross-entropy loss against the withheld true labels.
Training proceeds in two phases totaling
$K_{\mathrm{tot}} = 100{,}000$ gradient steps, completed in approximately 15 hours on a single node
with two NVIDIA H100 GPUs using data-parallel distributed training.

\paragraph{Phase 1: data curriculum.}
The first phase runs for $K = 100$ stages of $750$ steps each ($75{,}000$ steps total).
Within each stage, the SCM architecture is fixed to a randomly drawn configuration:
the number of hidden layers is sampled uniformly from $\{4,\ldots,12\}$ and the
hidden width from $\{12,\ldots,36\}$. A separate set of factors is resampled
independently for every batch, including noise level $\sigma_{\mathrm{noise}} \in \{0.005, 0.01,
0.02\}$, input distribution (Gaussian or uniform), activation schedule,
feature-source mode, and feature ordering, providing broad coverage of the prior
support regardless of stage. The number of input features per dataset is drawn
uniformly from $\{5,\ldots,20\}$ and the number of labeled positives from
$\{100,\ldots,300\}$, both sampled fresh every step.

The stage index controls the difficulty of the PU composition. Two quantities are
annealed linearly from stage $1$ to stage $K = 100$:
\begin{itemize}
    \item \textbf{Unlabeled-to-positive ratio} $\eta = n_u/P$: begins
    fixed at $\eta = 1$ and by the final stage is drawn uniformly from $[0.5,\,2.0]$,
    exposing the model to datasets with varying balance between the labeled and
    unlabeled pools.
    \item \textbf{Negative prevalence} $\pi$: begins fixed at $0.5$ and by the
    final stage is drawn uniformly from $[0.1,\, 0.9]$, training the model to
    handle the full spectrum from nearly pure-positive to nearly pure-negative
    unlabeled sets.
\end{itemize}
In addition, the probability that the generating SCM is causal increases linearly
with the stage index $s$ as $P(\mathrm{causal}) = s/(2K)$, reaching $0.5$ at the
last stage. Early training is therefore dominated by the simpler non-causal regime,
with causal datasets introduced gradually.

\paragraph{Phase 2: final-stage tail.}
After the curriculum, training continues for an additional $25{,}000$ steps with
all PU-composition parameters fixed at their final-stage ranges ($\eta \sim
\mathcal{U}[0.5, 2.0]$, $\pi \sim \mathcal{U}[0.1, 0.9]$). The learning rate is
scaled down by a factor of $4$ relative to Phase~1 (peak $4\times10^{-5}$, floor
$4\times10^{-6}$, with a $2{,}000$-step linear warmup). This tail phase allows the
model to consolidate representations under the hardest distribution without further
curriculum drift.

\paragraph{Optimizer and learning rate schedule.}
Both phases use the AdamW optimizer with
$\beta_1{=}0.9$, $\beta_2{=}0.95$, and weight decay $10^{-4}$. Gradients are
clipped to unit $\ell_2$ norm. In Phase~1, the learning rate follows a linear
warmup over the first $4{,}000$ steps to a peak of $1.6\times10^{-4}$, then decays
polynomially with exponent $1.5$ to a floor of $1.6\times10^{-5}$. An exponential
moving average of the model weights with decay $0.95$ is maintained throughout both
phases and used for all evaluation and inference.

\subsection{Inference}

At test time the model receives a PU dataset consisting of $P$ labeled positives and
$n_u$ unlabeled samples, and produces a probability estimate
$\hat{P}(y{=}+\mid\mathbf{x})$ for each unlabeled sample in a single forward pass.
No gradient computation, fine-tuning, or dataset-specific optimization is performed;
the model generalizes purely through the in-context mechanism learned during
pretraining.

\section{Experiments}
\label{sec:experiments}

In this section, we investigate the performance of the proposed PUICL method \footnote{\footnotesize An implementation of PUICL is publicly available at \url{https://anonymous.4open.science/r/puicl-58B1}.} on PU classification.
For comprehensive evaluations, we first describe several baseline methods and then discuss the empirical results.

\subsection{Settings}

\paragraph{Baseline methods.} We compare our proposed PUICL with four baseline methods:
\textbf{EMPU}~\citep{liu2024positive}, \textbf{VPU}~\citep{chen_variational_2020}, unconstrained least-squares importance fitting (\textbf{uLSIF})~\citep{hido_statistical_2011}, and \textbf{$\mathbf{(TED)}^{\mathbf n}$}~\citep{garg_mixture_2021}.
\begin{itemize}[leftmargin=*]
\item \textbf{EMPU}~\citep{liu2024positive} is a semiparametric PU learning method that formulates PU learning through density-ratio modeling and estimates the model parameters using an expectation-maximization (EM) algorithm, providing a closely related baseline for density-ratio-based PU learning.

\item \textbf{VPU}~\citep{chen_variational_2020} is a class-prior-free variational PU learning method that characterizes the gap between a given classifier and the Bayes classifier.

\item \textbf{uLSIF}~\citep{hido_statistical_2011} is a kernel-based density-ratio estimation method that directly estimates the importance weights between two distributions. Since uLSIF is not specifically designed for PU learning, we adapt it as a density-ratio baseline for PU classification.

\item \textbf{$\mathbf{(TED)}^{\mathbf n}$}~\citep{garg_mixture_2021} is an iterative transform-estimate-discard framework for PU learning. This method combines best-bin estimation of the negative prevalence $\pi$ with the conditional value ignoring risk (CVIR) objective and serves as a recent mixture-based PU learning baseline.
\end{itemize}


\paragraph{Data synthesis and models.} A total of 20 semi-synthetic PU classification tasks are constructed from real-world binary tabular datasets, including benchmark datasets from the UCI Machine Learning Repository, OpenML, and scikit-learn.
After preprocessing each dataset, the majority class is treated as the positive class.
To control the task size, at most 600 positive observations are sampled and then split into labeled and unlabeled positives in a 1:2 ratio, so that approximately one third of the positives are labeled and the remaining two thirds are unlabeled.
Negative observations are then added to the unlabeled pool to target a negative prevalence of $\pi = 0.5$; equivalently, when sufficient negatives are available, the number of unlabeled negatives is matched to the number of unlabeled positives.
Thus, when the positive cap is reached, each PU task contains approximately 200 labeled positives, 400 unlabeled positives, and 400 unlabeled negatives, while smaller datasets are adjusted according to the available numbers of positive and negative observations.

\paragraph{Evaluation metrics.}  To evaluate the performance,  for each synthetic dataset, we evaluate all methods using three standard metrics: area under the receiver operating characteristic curve (AUC), classification accuracy (Acc), and $F_1$-score. 
 AUC assesses ranking performance without requiring a fixed classification cutoff, whereas Acc and $F_1$-score evaluate the resulting class-label predictions. 
To reduce the effect of random sampling variability, the data generation and evaluation procedure is repeated 10 times.
\paragraph{Ranking metric: AUC.}
A score function $r(\mathbf{x}) := \hat P(y = +\mid \mathbf{x})$ can be obtained from PUICL and all four baselines, and is used to compute the empirical AUC via the Mann--Whitney U statistic:
$$
\widehat{\mathrm{AUC}} = \frac{1}{n_{u}^{(-)}\, n_{u}^{(+)}}\sum_{i:\, y_{u,i} = +}\sum_{j:\, y_{u,j} = -} \big[\, I\{r(\mathbf{x}_i) > r(\mathbf{x}_j)\} + \tfrac{1}{2}\,I\{r(\mathbf{x}_i) = r(\mathbf{x}_j)\}\,\big],
$$
where $n_{u}^{(+)}$ and $n_{u}^{(-)}$ are the numbers of positive and negative units in the unlabeled set, respectively, and $y_{u,i}$ is the true label of the $i$-th unlabeled observation (revealed only at evaluation).
\paragraph{Classification metrics: Acc and $F_1$-score.}

We further evaluate classification performance using Acc and the $F_1$-score. 
Acc is defined as the proportion of correctly classified observations in the test set. 
The $F_1$-score, defined as the harmonic mean of precision and recall, summarizes the trade-off between these two quantities, where precision is the proportion of predicted positives that are truly positive, and recall is the proportion of true positives that are correctly identified.
For PU data, the $F_1$-score emphasizes the ability to recover the positive class, whereas Acc measures overall classification accuracy.

\subsection{Results and Analysis}

The empirical results on the 20 semi-synthetic PU classification tasks are summarized in Table~\ref{tab:result080-all}. 
The table reports the AUC, accuracy (Acc), and $F_1$-score for all competing methods, with the highest value for each dataset and each metric highlighted in bold.

In terms of AUC, PUICL consistently demonstrates strong performance among the five competing methods. 
It achieves the highest AUC on the majority of datasets and also attains the largest cross-dataset average AUC (0.882), as shown in the last row. 
Among the baseline methods, uLSIF is the most competitive in average AUC, with an average value of 0.841, followed by $\text{(TED)}^{n}$ and EMPU with average AUC values of 0.832 and 0.824, respectively.
VPU obtains a lower average AUC of 0.752, although it performs strongly on some individual datasets such as MONK-1.

For classification performance, Acc and $F_1$-score are reported for the four methods that provide explicit class predictions. 
PUICL achieves the best average Acc (0.802) and attains the highest Acc on most datasets. 
For the $F_1$-score, PUICL remains highly competitive, with an average value of 0.758, which is close to the best average value of 0.759 achieved by $\text{(TED)}^{n}$. 
Notably, PUICL achieves the highest $F_1$-score on the majority of datasets, indicating strong performance in identifying the positive class. 
Overall, these results demonstrate that PUICL delivers robust and well-balanced performance across different evaluation metrics, combining strong ranking ability with competitive classification accuracy in PU learning.

\begin{table}[htbp]
\centering
\scriptsize
\setlength{\tabcolsep}{2.5pt}
\caption{Average AUC, accuracy (Acc), and $F_1$-score of competing methods on 20 semi-synthetic PU classification tasks. The best result for each dataset and metric is highlighted in \textbf{bold}. PUICL consistently achieves strong performance across datasets, particularly in terms of AUC and accuracy. }
\label{tab:result080-all}
\renewcommand{\arraystretch}{1.1}
\begin{tabular}{lccccc|cccc|cccc}
\toprule
& \multicolumn{5}{c|}{AUC} 
& \multicolumn{4}{c|}{Acc} 
& \multicolumn{4}{c}{$F_1$-score} \\
\cmidrule(lr){2-6} \cmidrule(lr){7-10} \cmidrule(lr){11-14}
Dataset 
& PUICL & EMPU & VPU & uLSIF & $\text{(TED)}^{n}$ 
& PUICL & EMPU & VPU & $\text{(TED)}^{n}$ 
& PUICL & EMPU & VPU & $\text{(TED)}^{n}$ \\
\midrule
Diabetes & \textbf{0.809} & 0.789 & 0.782 & 0.786 & 0.759 & \textbf{0.714} & 0.666 & 0.670 & 0.690 & 0.690 & 0.677 & 0.586 & \textbf{0.694} \\
Haberman & \textbf{0.685} & 0.674 & 0.612 & 0.666 & 0.649 & \textbf{0.682} & 0.527 & 0.673 & 0.645 & 0.395 & 0.452 & 0.290 & \textbf{0.473} \\
Heart & \textbf{0.871} & 0.842 & 0.732 & 0.862 & 0.846 & 0.759 & 0.722 & 0.655 & \textbf{0.774} & \textbf{0.784} & 0.757 & 0.609 & 0.779 \\
ILPD & \textbf{0.723} & 0.696 & 0.667 & 0.619 & 0.610 & \textbf{0.645} & 0.576 & 0.618 & 0.596 & 0.419 & \textbf{0.586} & 0.038 & 0.461 \\
MONK-1 & 0.846 & 0.743 & \textbf{0.895} & 0.813 & 0.796 & 0.727 & 0.726 & \textbf{0.772} & 0.720 & 0.729 & 0.663 & \textbf{0.790} & 0.723 \\
MONK-2 & \textbf{0.824} & 0.506 & 0.600 & 0.638 & 0.625 & \textbf{0.671} & 0.539 & 0.558 & 0.595 & 0.489 & 0.269 & 0.482 & \textbf{0.559} \\
MONK-3 & \textbf{0.975} & 0.846 & 0.936 & 0.942 & 0.955 & 0.780 & 0.718 & 0.782 & \textbf{0.901} & 0.823 & 0.750 & 0.818 & \textbf{0.904} \\
Abalone & \textbf{0.900} & 0.900 & 0.876 & 0.879 & 0.882 & 0.810 & \textbf{0.811} & 0.790 & 0.801 & 0.802 & \textbf{0.810} & 0.773 & 0.796 \\
Adult & \textbf{0.855} & 0.820 & 0.500 & 0.826 & 0.796 & \textbf{0.762} & 0.721 & 0.500 & 0.733 & \textbf{0.756} & 0.732 & 0.000 & 0.732 \\
Banknote & \textbf{1.000} & 0.999 & 1.000 & 1.000 & 1.000 & 0.989 & 0.980 & 0.980 & \textbf{0.993} & 0.989 & 0.980 & 0.981 & \textbf{0.993} \\
Car & \textbf{0.919} & 0.735 & 0.796 & 0.784 & 0.794 & \textbf{0.846} & 0.702 & 0.763 & 0.791 & \textbf{0.642} & 0.403 & 0.482 & 0.579 \\
Default Credit & \textbf{0.724} & 0.663 & 0.500 & 0.703 & 0.657 & \textbf{0.652} & 0.610 & 0.500 & 0.629 & 0.616 & 0.579 & 0.000 & \textbf{0.620} \\
Iranian Churn & \textbf{0.952} & 0.904 & 0.667 & 0.909 & 0.899 & \textbf{0.857} & 0.778 & 0.577 & 0.832 & \textbf{0.867} & 0.812 & 0.290 & 0.831 \\
Letter C vs U & \textbf{0.999} & 0.992 & 0.993 & 0.991 & 0.996 & \textbf{0.977} & 0.925 & 0.942 & 0.974 & \textbf{0.978} & 0.930 & 0.945 & 0.974 \\
MAGIC Gamma & \textbf{0.897} & 0.799 & 0.795 & 0.870 & 0.847 & \textbf{0.808} & 0.710 & 0.701 & 0.773 & \textbf{0.808} & 0.671 & 0.698 & 0.771 \\
Mushroom & \textbf{0.993} & 0.959 & 0.952 & 0.968 & 0.962 & \textbf{0.966} & 0.873 & 0.857 & 0.940 & \textbf{0.967} & 0.887 & 0.864 & 0.940 \\
Rice & 0.975 & \textbf{0.978} & 0.500 & 0.970 & 0.966 & 0.913 & \textbf{0.921} & 0.500 & 0.911 & 0.910 & \textbf{0.922} & 0.000 & 0.910 \\
Spambase & \textbf{0.913} & 0.888 & 0.745 & 0.856 & 0.889 & \textbf{0.819} & 0.786 & 0.671 & 0.806 & \textbf{0.835} & 0.811 & 0.649 & 0.826 \\
WDBC & \textbf{0.994} & 0.976 & 0.900 & 0.978 & 0.987 & \textbf{0.955} & 0.775 & 0.839 & 0.950 & \textbf{0.953} & 0.820 & 0.783 & 0.947 \\
Wine Quality & \textbf{0.780} & 0.764 & 0.586 & 0.758 & 0.729 & \textbf{0.702} & 0.657 & 0.531 & 0.674 & \textbf{0.710} & 0.648 & 0.254 & 0.673 \\
\hline
\textbf{Average} 
& \textbf{0.882} & 0.824 & 0.752 & 0.841 & 0.832
& \textbf{0.802} & 0.736 & 0.694 & 0.786
& 0.758 & 0.708 & 0.517 & \textbf{0.759} \\
\bottomrule
\end{tabular}
\end{table}
\section{Discussion}
\label{sec:discussion}
We introduced PUICL, a transformer-based model that solves PU classification entirely through in-context learning. Pretrained on synthetic datasets generated from randomly instantiated structural causal models, PUICL takes a labeled-positive set and an unlabeled pool as a single input and returns class probabilities in one forward pass, with no gradient updates or per-task fitting. PUICL outperforms four standard PU learning baselines in average AUC and accuracy and is competitive on $F_1$-score, while remaining compact at roughly only 1.23M trainable parameters.

Several limitations should be acknowledged. First, the prior generation procedure encodes the SCAR assumption, so PUICL inherits the misspecification that affects most SCAR-based PU methods when the labeling mechanism is informative. Second, the model is trained with at most 20 features, a few hundred labeled positives, and class priors restricted to $\pi \in [0.1, 0.9]$; performance outside this regime---particularly for high-dimensional data or extreme class imbalance---is not guaranteed. Third, although PUICL implicitly reasons about the mixture composition of the unlabeled pool, it does not return an explicit estimate of $\pi$, which limits its use in applications requiring calibrated decision thresholds or interpretable class priors.
These limitations suggest several directions for future work. The prior could be extended to incorporate explicit selection mechanisms, enabling PUICL to handle non-SCAR data.
Architecture and prior extensions would enable scaling to larger feature dimensions and sample sizes, while augmenting the output head with a class-prior estimator and calibration module would yield interpretable $\pi$ estimates. Finally, the in-context formulation extends naturally to other forms of weak supervision, suggesting that PUICL is a first step toward a broader family of in-context semi-supervised tabular learners.

\bibliographystyle{plainnat}
\bibliography{references.bib}

\end{document}